\documentclass[letterpaper, 10 pt, conference]{ieeeconf}  

\IEEEoverridecommandlockouts                              

\usepackage{cite}
\usepackage{amsmath,amssymb,amsfonts}
\usepackage[graphicx]{realboxes}
\usepackage{textcomp}
\usepackage[table,xcdraw]{xcolor}
\usepackage{multirow}
\usepackage{array}
\usepackage{tikz}
\usepackage{siunitx}
\usepackage{lscape}
\usepackage{adjustbox}
\usepackage{tabularx}
\usepackage{mleftright}
\usepackage{etoolbox}
\usepackage{xcolor}
\usepackage{amsmath}
\usepackage{amssymb}
\usepackage[normalem]{ulem}
\usepackage{tablefootnote}
\usepackage{verbatim}
\usepackage{graphicx}
\usepackage{booktabs}
\usepackage{mathtools}
\usepackage{makecell}

\usepackage{algorithm}
\usepackage{algpseudocode}

\algtext*{EndFor}   
\algtext*{EndIf}    
\algtext*{EndWhile} 
\algtext*{EndFunction} 

\title{\LARGE \bf
ID-EA: Identity-driven Text Enhancement and Adaptation \\ with Textual Inversion for Personalized Text-to-Image Generation
}
\author{Hyun-Jun Jin$^{1}$, Young-Eun Kim$^{1}$, and Seong-Whan Lee$^{1}$
\thanks{This research was supported by the Institute of Information \& Communications Technology Planning \& Evaluation (IITP) grant, funded by the Korea government (MSIT) (No. RS-2019- II190079 (Artificial Intelligence Graduate School Program (Korea University)), and No. RS-2024-00457882 (AI Research Hub Project)).}
\thanks{$^{1}$H.-J Jin, Y.-E. Kim, and S.-W. Lee are with the Department of Artificial Intelligence, Korea University, Anam-dong, Seongbuk-ku, Seoul 02841, Korea.
    {\tt\small \{hyunjun\_jin, ye\_kim, sw.lee\}@korea.ac.kr}
}
}
\begin{document}

\maketitle
\thispagestyle{empty}
\pagestyle{empty}

\begin{abstract}
Recently, personalized portrait generation with a text-to-image diffusion model has significantly advanced with Textual Inversion, emerging as a promising approach for creating high-fidelity personalized images. Despite its potential, current Textual Inversion methods struggle to maintain consistent facial identity due to semantic misalignments between textual and visual embedding spaces regarding identity. We introduce ID-EA, a novel framework that guides text embeddings to align with visual identity embeddings, thereby improving identity preservation in a personalized generation. ID-EA comprises two key components: the ID-driven Enhancer (ID-Enhancer) and the ID-conditioned Adapter (ID-Adapter). First, the ID-Enhancer integrates identity embeddings with a textual ID anchor, refining visual identity embeddings derived from a face recognition model using representative text embeddings. Then, the ID-Adapter leverages the identity-enhanced embedding to adapt the text condition, ensuring identity preservation by adjusting the cross-attention module in the pre-trained UNet model. This process encourages the text features to find the most related visual clues across the foreground snippets. Extensive quantitative and qualitative evaluations demonstrate that ID-EA substantially outperforms state-of-the-art methods in identity preservation metrics while achieving remarkable computational efficiency, generating personalized portraits approximately 15 times faster than existing approaches. 
\end{abstract}

\section{INTRODUCTION}

Recent advancements in large-scale diffusion models \cite{saharia2022photorealistic,ramesh2022hierarchical,rombach2022high} have enabled the generation of realistic images from natural language descriptions. These text-to-image generation frameworks have significantly expanded the range of generative tasks, particularly through text-to-image personalization methods\cite{gal2022image,hertz2022prompt,balaji2022ediff,ruiz2023dreambooth,fujisawa1999information,yang2007reconstruction}. Personalization methods \cite{gal2022image,ruiz2023dreambooth} leverage a few reference images representing a specific target concept, allowing users to create personalized images within a diverse range of novel contexts or distinctive styles. Among these methods, Textual Inversion \cite{gal2022image} is notable for its capability to learn target concepts by converting reference images into corresponding textual embeddings. By training these target concepts and optionally fine-tuning selected components of the model, the approach facilitates the generation of novel depictions or variations of the same subject using textual prompts alone.

\begin{figure}[t!] 
  \centerline{\includegraphics[width=0.48\textwidth]{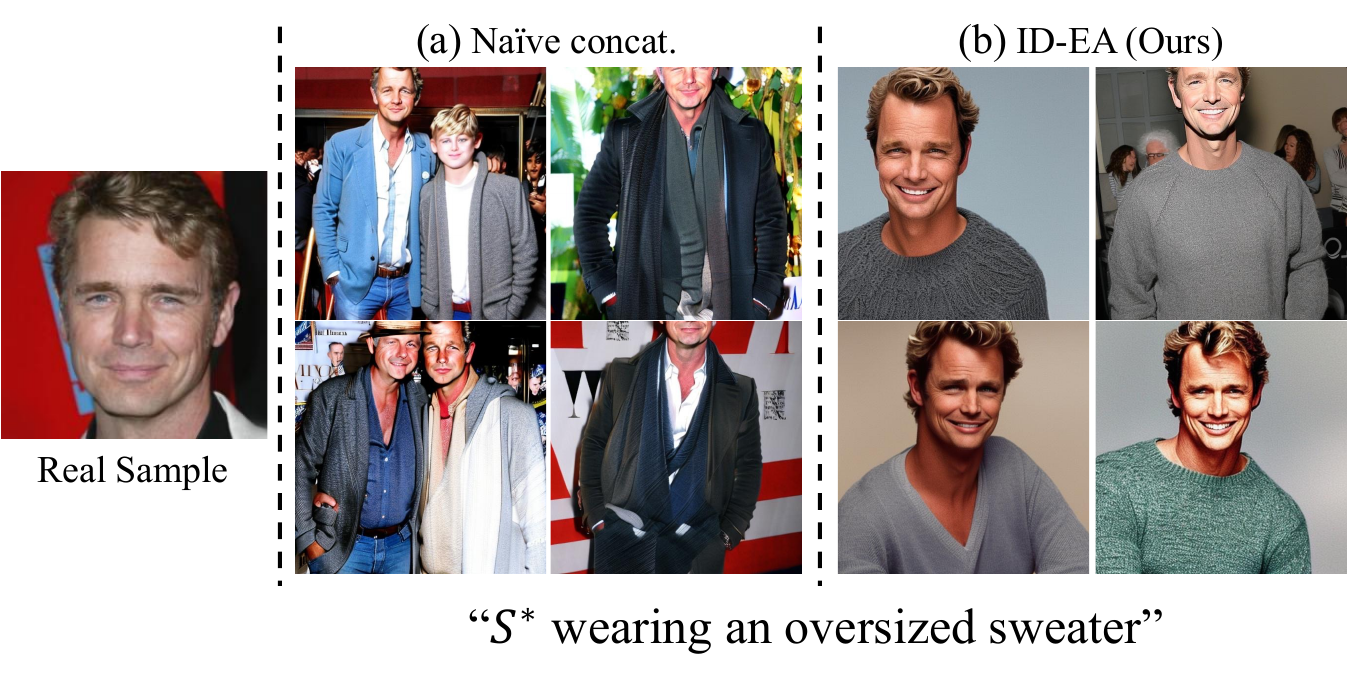}}
  \caption{
The comparison of text fidelity between the previous identity enhancement method using FRNet and our proposed method. The previous method using FRNet was generated by concatenating the FRNet embedding \( E_f \) and the target concept \( c_{\theta} \). In our method (b), the results with the ID-EA are presented.
  } 
  \label{fig1} 
\end{figure}

\begin{figure}[t!] 
  \centerline{\includegraphics[width=0.48\textwidth]{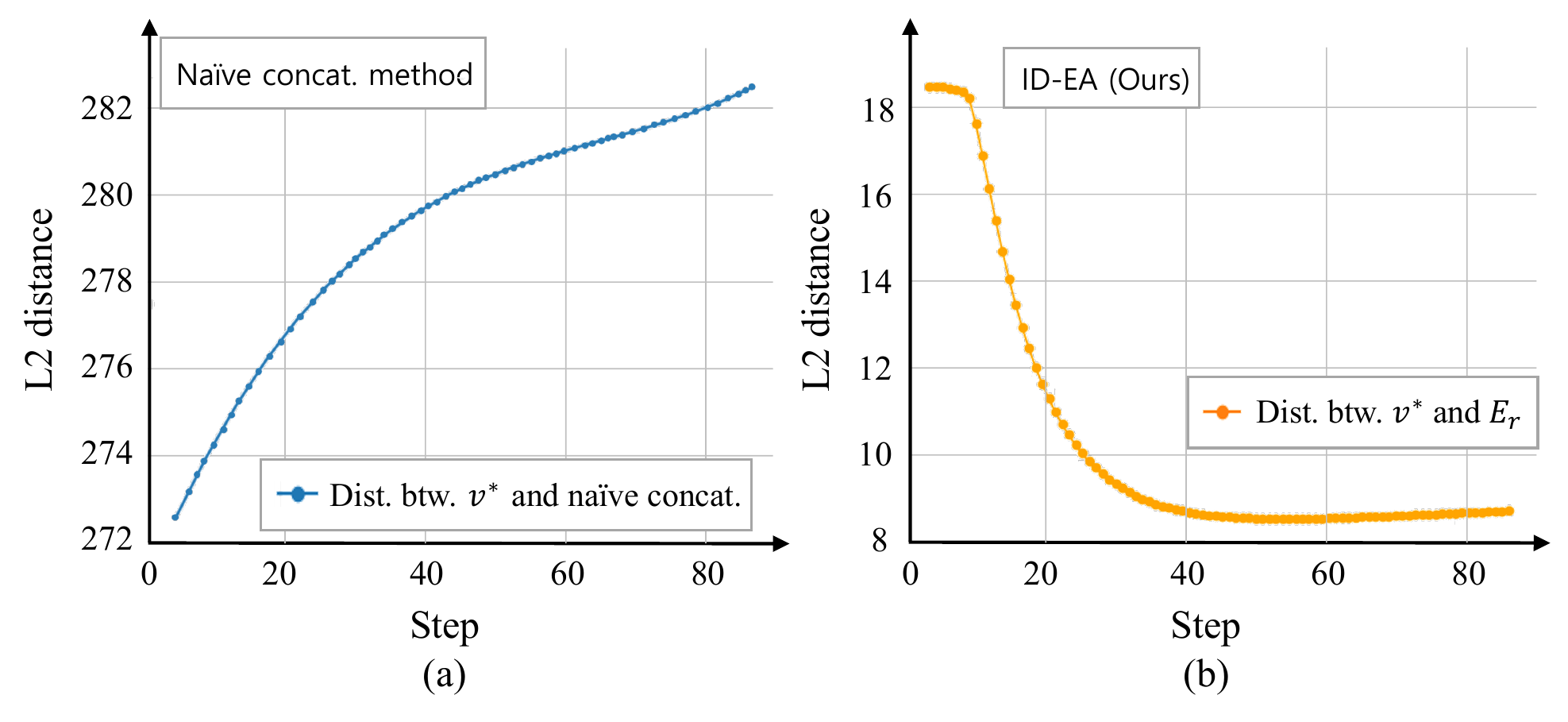}}
  \caption{
Comparison of L2 distances between the target text embedding \(v^*\) and combined embeddings. Fig. (a) shows naive concatenation of \(c_\theta\) and \(E_f\), which fails to align with \(v^*\), showing a large gap from target text embedding. Fig. (b) uses identity-enhanced embedding \(E_r\), where the proposed ID-Enhancer helps reduce this gap, leading to the face space closer to the textual embedding space. Both use a trained linear layer for \(E_f\).
}

  \label{fig2} 
\end{figure}

A common challenge with these methods is that they emphasize auxiliary elements rather than subjects, resulting in inappropriate preservation of facial identities. To address this issue, Celeb Basis \cite{yuan2023inserting} integrates a face recognition network (FRNet), such as ArcFace \cite{deng2019arcface}, to enhance identity representation. Despite the integration of FRNet to capture identity information, resulting in irrelevant prompt representations, as shown in Fig. \ref{fig1}(a). Our investigation reveals that, as shown in Fig.~\ref{fig2}(a), this approach suffers from an L2 distance discrepancy between the target text and image embeddings. This misalignment issue, incorporating identity features without adequately aligning text and image modalities, leads to identity fidelity and overall generative consistency.

To address this issue, our approach aims to mitigate the inconsistency between text and image embeddings and introduces an adapter specifically designed to augment facial parameters. We introduce Identity-driven Enhancement and Adaptation (ID-EA), a novel framework designed to align latent embedding spaces by integrating identity embeddings from FRNet with textual embeddings, thus enhancing identity preservation in personalized text-to-image generation. Our method consists of two distinct modules. Our ID-driven Enhancer (ID-Enhancer) module synthesizes features from FRNet embedding and textual ID anchor of a person during training, mitigating overfitting to non-identity attributes. Furthermore, we propose an ID-conditioned Adapter (ID-Adapter), which leverages identity-enhanced embedding to adapt to text conditions, facilitating a personalization method that is both lightweight and effective. ID-Adapter progressively integrates image information into the learned ID-conditioned embedding, associating image features with textual embeddings, as illustrated in Fig.~\ref{fig1}(b) and Fig.~\ref{fig2}(b). We demonstrate the superior performance of ID-EA compared to the baseline methods\cite{gal2022image,lee2015motion,ruiz2023dreambooth,yuan2023inserting,guo2024pulid} through qualitative and quantitative evaluations. 

Our method enables a variety of personalized face generations with high visual fidelity and less training time.

\section{RELATED WORKS}

\subsection{Text-guided Synthesis}

Recently, the diffusion model has emerged as a leading method for text-based synthesis and has shown impressive capabilities in generating detailed and realistic images. Models such as Imagen\cite{saharia2022photorealistic}, DALL-E2\cite{ramesh2022hierarchical}, and Stable Diffusion\cite{rombach2022high} utilize text encoder (e.g., CLIP\cite{radford2021learning}) to bridge the gap between textual descriptions and visual representations. Although these models excel in general image synthesis tasks, their performance decreases in scenarios requiring detailed personalization task\cite{balaji2022ediff,hyung2024magicapture,wu2023singleinsert,lee2001automatic,lee1995multilayer}. This limitation arises from the CLIP text encoder's struggle to fully capture specific subject details. To address this issue, we propose an identity enhancement and prompt adaptation method aimed at refining both textual and visual feature representations, thus facilitating the synthesis of high-quality images.

\subsection{Text-to-image Personalization Methods}
Personalization in text-to-image synthesis methods\cite{wu2023singleinsert,gal2022image,ruiz2023dreambooth,lee2024text} involves adapting pre-trained generative models to create images of specific subjects or concepts. SingleInsert\cite{wu2023singleinsert} into pre-trained models by either fine-tuning the entire model, optimizing embeddings, or adapting a subset of model parameters. Recent methods~\cite{yuan2023inserting,peng2024portraitbooth} integrate external face recognition networks (FRNet) into the generation process to better capture fine-grained identity features, rather than relying solely on CLIP-based text embeddings. By optimizing an identity similarity alongside the diffusion objective, PortraitBooth~\cite{peng2024portraitbooth} achieves improved consistency between the generated images and the subject’s true appearance. Celeb Basis~\cite{yuan2023inserting} effectively optimizes target text embeddings through a designed basis in the textual space and further integrates an FRNet to enhance identity preservation. Despite notable successes, these methods have challenges in the modality mismatch, which refers to the inconsistency between textual embedding space and visual representation for the target subject. Our ID-EA introduces an identity embedding alignment mechanism that explicitly bridges the gap between textual and visual representations of a given subject.


\section{METHOD}
\subsection{Preliminaries}
We implement our training approach based on the Stable Diffusion Model (SDM) \cite{rombach2022high}, which synthesizes high-quality images within a compressed latent space through diffusion processes. These models employ a DDPM \cite{ho2020denoising} to generate latent representations.
In text-to-image generation tasks, the process is conducted with textual descriptions provided as input prompts. Given a textual prompt \( y \), the input sentence is tokenized into \( n \) discrete tokens. Each token is mapped to a text embedding \( v_i \in \mathbb{R}^{1024} \) via a fixed embedding lookup table. The resulting sequence of embeddings \( [v_1, \dots, v_n] \) is processed through a pre-trained CLIP text encoder \cite{radford2021learning}, producing a conditional embedding vector \( c(y) \). This vector guides the diffusion model in generating latent codes that accurately reflect the semantic content of the prompt. To further personalize this generation process, we integrate the Textual Inversion method \cite{alaluf2023neural} by introducing an identity-specific token \( S^* \), representing a novel target concept. The embedding corresponding to this identity-specific token, \( v^* \), is directly optimized within the diffusion model by minimizing the following loss function:
\begin{equation}
v^{*} = \arg\min_{v} \mathbb{E}{z, y, \varepsilon, t}\left[|\varepsilon - \varepsilon{_\theta}(z_{t}, t, c(y, v))|_{2}^{2}\right],
\end{equation}
where \( t \) denotes the timestep, \( z_{t} \) represents the latent variable corrupted with noise, and \( c(y, v) \) denotes the conditioning vector derived from the prompt \( y \) and the textual embedding \( \textit{}v \), the denoising network \( \varepsilon_{\theta}(\cdot) \) aim to remove the noise added to the original latent code \( z_{0} \). Consequently, optimizing the textual embedding \( v^{*} \) minimizes the SDM loss.

\begin{figure*}[t] 
  \centerline{\includegraphics[width=0.96\textwidth]{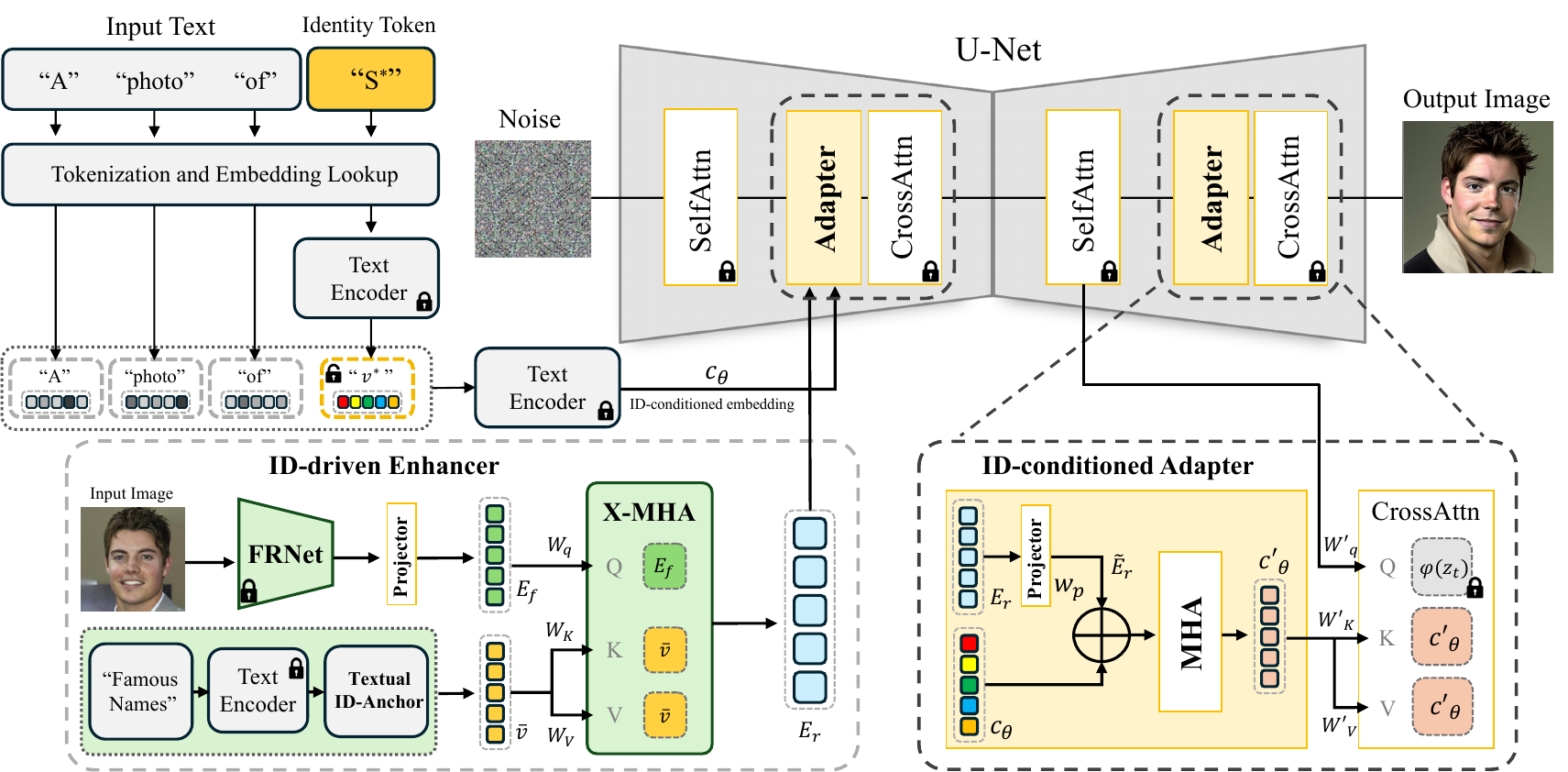}}
  \caption{
Overview of our proposed ID-EA. The unique identifier \( v^* \) is extracted from the text encoder, serving as the basis for embedding initialization. Subsequently, the identity-enhanced embedding $E_r$ is generated through the ID-driven Enhancer (ID-Enhancer) method by integrating embeddings obtained from the textual ID anchor and FRNet. The resulting $E_r$ acts as a conditioning vector for the UNet, guided by the output vector \( c_{\theta} \) and the ID-conditioned Adapter (ID-Adapter), and is utilized within the Cross-Attention mechanism as the key and value inputs.
  } 
  \label{fig3} 
\end{figure*}

\subsection{ID-driven Enhancer}

Our goal is to enhance identity preservation by integrating textual and visual identity representations while reducing discrepancies between their respective embedding spaces. To this end, we propose ID-driven Enhancer (ID-Enhancer), which refines visual identity embeddings extracted from a pre-trained FRNet (e.g., ArcFace \cite{deng2019arcface}) using a textual ID anchor. Conditioned on the textual ID anchor derived from person names, ID-Enhancer generates identity-enhanced embeddings that more faithfully capture the subject's identity and facilitate improved alignment with corresponding text prompts during personalized text-to-image generation.

We first construct a textual ID anchor by defining an embedding set ${U}=\{v_j\}_{j=1}^{m}$, utilizing $m=691$ widely recognized names, following \cite{yuan2023inserting}. We calculated as: $\bar{v} = \frac{1}{m} \sum_{j=1}^{m} v_j.$ Each name is represented by two token embeddings, corresponding separately to the first and last names. The textual ID anchor is represented as 
$\bar{v} =( \bar{v}{_f}, \bar{v}{_l}),$
where \(\bar{v}{_f}\) and \(\bar{v}{_l}\) denote the embeddings for first and last names, respectively.

Given a reference image and a text prompt, we extract the initial image features $E_f = \phi_{\text{img}}(X) \in \mathbb{R}^{M \times d}$ using a FRNet $\phi_{\text{img}}(X)$ (e.g.,  ArcFace~\cite{deng2019arcface}). Here, $d$ denotes the dimensionality of the embeddings, and $M$ represents the number of visual tokens. $N$ indicates the number of textual tokens.
When the image features serve as the
query and the text features as the key and value, the operation is
formulated as follows:

\begin{figure*}[t] 
  \centerline{\includegraphics[width=1\textwidth]{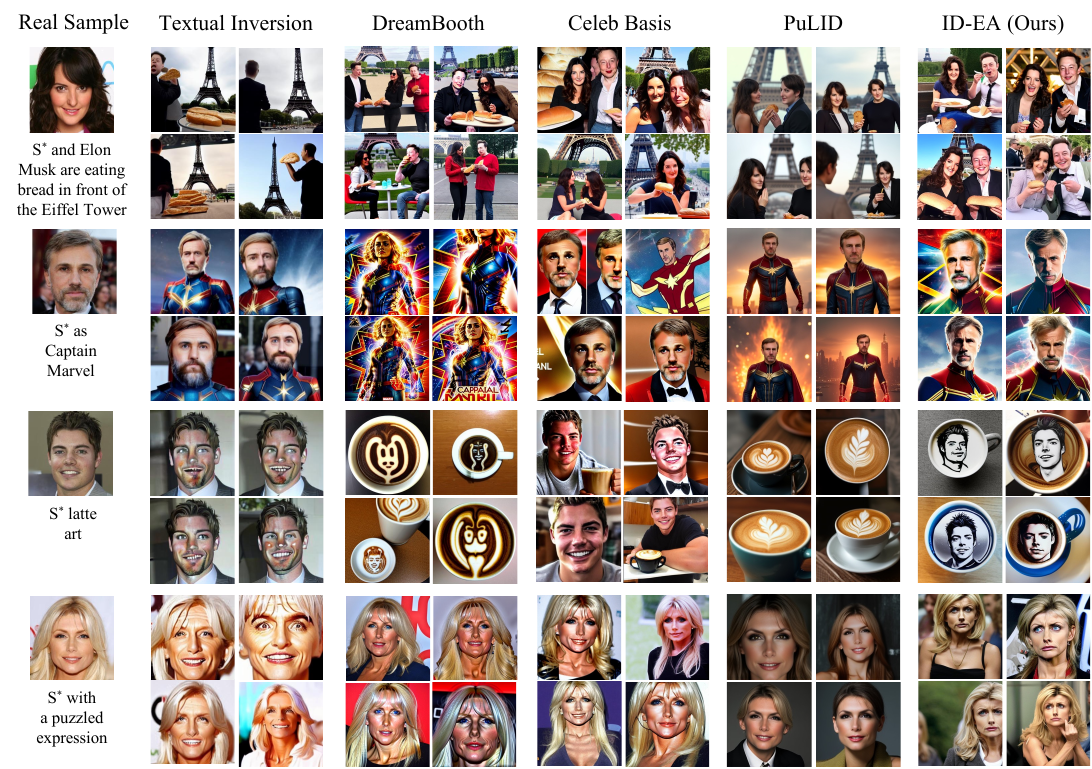}}
  \caption{
  Qualitative comparison with previous methods. Given an input image, we show four images generated by each method using the same random seed. Our approach demonstrates improved identity clarity and enhanced text-driven editing capabilities compared to existing approaches, as evidenced by experiments across diverse tasks such as simultaneous generation of multiple subjects and varied backgrounds (e.g., \( S^* \) and Elon Musk eating bread in front of the Eiffel Tower), conceptual embedding of input subjects (e.g., \( S^* \) as Captain Marvel), stylistic transformation of subjects into novel artistic forms (e.g., \( S^* \) latte art), and facial expression editing (e.g., \( S^* \) with a puzzled expression). These advancements validate the significant contributions of our proposed ID-Enhancer and ID-Adapter modules in preserving subject identity and facilitating precise text-driven image manipulation.
  } 
  \label{fig4} 
\end{figure*}

\begin{equation}
\begin{split}
E_{r} &= \text{X-MHA}(E_{f}, \bar{v}) \\
&= \text{Softmax} \left( \frac{\mathbf{Q}(E_{f})\mathbf{K}(\bar{v})^{\top}}{\sqrt{d}} \right) \mathbf{V}(\bar{v}),
\end{split}
\end{equation}
where $\mathbf{Q} = W_q \cdot E_f$, $\mathbf{K}=W_k\cdot \bar{v}$, $\mathbf{V}=W_v \cdot \bar{v}$ are the query, key, and value matrices of the multi-head cross-attention (X-MHA) operation, respectively, and $W_q$, $W_k$, $W_v$ are the weight matrices of the trainable linear projection layers.

\subsection{ID-conditioned Adapter}
To incorporate the identity-enhanced embedding with the target concept, we propose an ID-conditioned Adapter (ID-Adapter). ID-Adapter modulates only the cross-attention modules within the frozen U-Net to inject identity cues, enabling a lightweight yet practical personalization approach.
Our approach implements the model by introducing trainable adapter layers, optimizing the key \(W'_K\) and values \(W'_V\) matrices within the cross attention block (CrossAttn).

Our ID-Adapter conducts a multi-head self-attention mechanism (MHA) between the ID-conditioned embeddings \(c_{\theta}\) and the identity-enhanced embedding \(E_r\). To achieve this, we first update the original embedding \(E_r\), with a linear transformation as: $\tilde{E}_r = W_p \cdot E_r$, where $W_p$ denotes the projection matrix, and $\tilde{E}_r$ represents the transformed embedding. These transformed embeddings are then concatenated with \(c_{\theta}\) to form the self-attention input: $\tilde{C} =  \text{Concat}\bigl(c_{\theta}, \tilde{E}_r\bigr),$
where \(c_{\theta}\in \mathbb{R}^{N \times d}\) denotes the CLIP text embedding produced by the text encoder, \(\tilde{E}_r\in \mathbb{R}^{M \times d}\) denotes the identity-enhanced embedding by the ID-Enhancer, and \(\tilde{C} \in \mathbb{R}^{(N+M) \times d}\) represents the combined embedding. Here, \(N\) and \(M\) represent the respective context lengths for the text and enhanced embeddings, while \(d\) is the dimensionality of \(E_r\).
Following the MHA on \(\tilde{C}\), we extract a subset of vectors \(C'\) that correspond to the target text tokens of length \(N\). Subsequently, the scaling factors \(\gamma\) and \(\beta\) are introduced to strengthen the overall learning and information transfer within the ID-Adapter.
The specific fusion between \(C'\) and the condition embedding is performed as follows:
\begin{equation}
c'_{\theta}  = c_{\theta} + \beta \cdot \tanh (\gamma )\cdot \text{MHA}(C'),
\end{equation}
where $c'_{\theta}$ denotes the output of the condition embedding, \( \gamma \) is a learnable scalar initialized at zero, and \( \beta \) is a constant balancing the significance of the adapter layer. 

The condition embedding \( c'_{\theta} \) is integrated into each CrossAttn, allowing image features to be conditioned on the textual embedding during training. Then, by updating the key and value projection matrices within each CrossAttn, the model can attend to visual facial attributes and associate them with personalized concepts within the textual embedding space. The image features are integrated into the pre-trained UNet model by the adapted modules with CrossAttn:

\begin{equation}
\begin{split}
\textbf{Z} &= \text{CrossAttn}(\varphi(z_t), c'_\theta) \\
&= \text{Softmax} \left( \frac{\mathbf{Q'}(\varphi(z_t))\mathbf{K'}(c'_\theta)^{\top}}{\sqrt{d}} \right) \mathbf{V'}(c'_\theta),
\end{split}
\end{equation}
where $\textbf{Z}$ is new CrossAttn of pre-trained U-Net. $\mathbf{Q'}=W'_q \cdot \varphi(z_t)$, $\mathbf{K}'=W'_k\cdot c'_\theta$, $\mathbf{V}'=W'_v\cdot c'_\theta$ are the query, key, and values matrices of the attention operation respectively. $\varphi(z_t)$ is the hidden states through the UNet implementation. We freeze the parameters of the original UNet model, making only the $W'_k$ and $W'_v$ trainable.
\begin{figure*}[t] 
  \centerline{\includegraphics[width=1\textwidth]{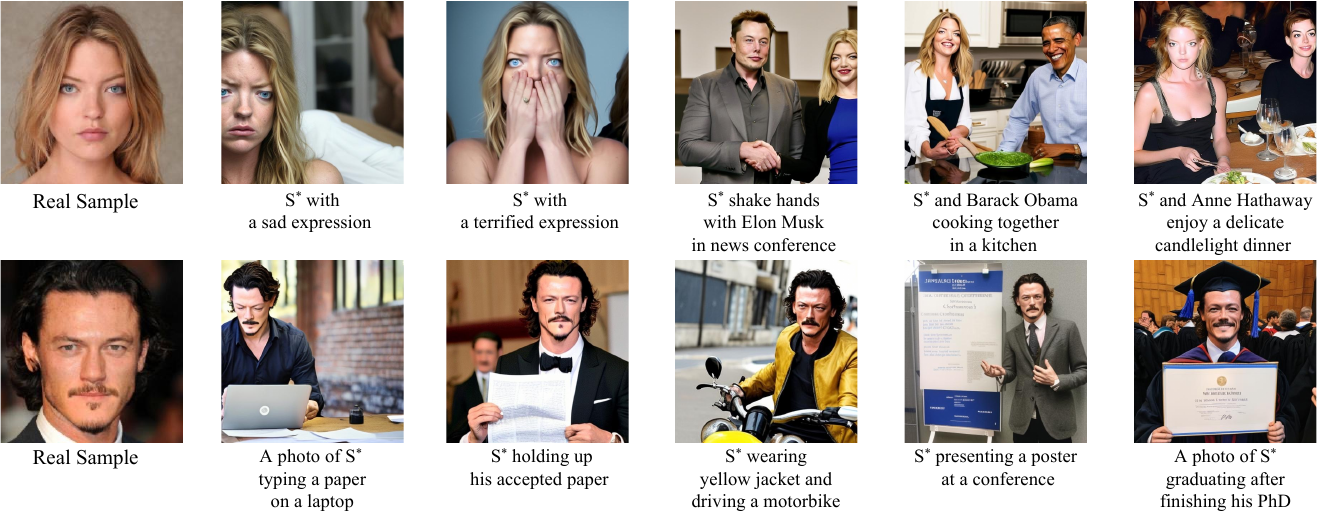}}
  \caption{
Examples of personalized text-to-image generation with ID-EA. Our method simultaneously generates diverse facial expressions for a single subject within varying backgrounds and interactions involving multiple individuals. It also allows easy textual editing of interactions between subjects and their backgrounds. Furthermore, our approach consistently maintains subject identity across diverse textual conditions.
  } 
  \label{fig5}
\end{figure*}


\section{EXPERIMENTS}

\subsection{Implementation Details}
We utilize the Stable Diffusion v2.1~\cite{rombach2022high} as our base model. All images are generated at a resolution of 512 × 512. A single image is provided as input, and the training process is conducted on a single NVIDIA A40 GPU with a batch size of 8 and a learning rate of 0.0015. All results are obtained using 300 optimization steps.

\subsection{Experiment Setup}
We evaluate each method using images from the CelebA-HQ test set\cite{liu2015deep} with prompts primarily sourced from \cite{yuan2023inserting}. Our method is compared against three state-of-the-art personalization techniques: Textual Inversion~\cite{gal2022image}, Dreambooth~\cite{ruiz2023dreambooth}, Celeb Basis~\cite{yuan2023inserting}, and PuLID~\cite{guo2024pulid}. All methods are implemented for one-shot personalization. For quantitative evaluation, each method is assessed using the first 50 images from the CelebA-HQ test set based on three metrics: identity similarity, prompt similarity, and image quality assessment (IQA). Identity preservation in generated images is measured using ArcFace\cite{deng2019arcface}, a pre-trained facial recognition model. The similarity between images and text prompts is calculated using the CLIP score, employing the metrics proposed in Celeb Basis\cite{yuan2023inserting}. Lastly, IQA utilizes CLIP-based image quality assessment\cite{wang2023exploring}.

\subsection{Experiment Results}
\textbf{Qulitative Evaluation.} 
Fig.~\ref{fig4} visually compares personalized image generation using four prompts: background modification, individual interaction, artistic style, and expression editing. 
Textual Inversion~\cite{gal2022image} exhibits an overfitting problem, failing to compose the given individual in novel scenes.
 DreamBooth~\cite{ruiz2023dreambooth} and Celeb Basis~\cite{yuan2023inserting} struggle with background modifications and the generation of images involving celebrities. In particular, when reconstructing an individual under complex editing prompts, such as diverse artistic styles, DreamBooth~\cite{ruiz2023dreambooth} and Celeb Basis~\cite{yuan2023inserting} tend to disregard the identity of the conditioned individual, prioritizing only prompt conditions. PuLID~\cite{guo2024pulid} struggles to generate background modification, artistic styles, and expression editing as instructed by the prompts, indicating a reliance on generalized image features rather than precise prompt-specific modifications. Compared to these methods,  our ID-EA consistently preserves the individual's identity while offering robust editing capabilities across all prompts. Fig.~\ref{fig5} presents the results of generating various prompts from a single reference image of a person. Specifically, the first row demonstrates expression editing and depicts the subject interacting with multiple individuals across various backgrounds. In addition, the second row illustrates the reference subject engaging in individual interactions within different contextual environments.
\begin{table}[t!]
\caption{Quantitative comparison in “person” category of conventional methods and our method. the metric ``Identity" denotes the identity similarity, ``Prompt" denotes the similarity between generated images and given text prompts, and ``IQA" denotes the image quality assessment. ``Time" indicates the average personalization duration in seconds.} 
    \centering
    \renewcommand{\arraystretch}{1.5} 
    \setlength{\tabcolsep}{5pt} 
    \begin{tabular}{lcccc}
        \hline
        \textbf{Methods} & \textbf{Identity $\uparrow$} & \textbf{Prompt $\uparrow$} & \textbf{IQA $\uparrow$} & \textbf{Time $\downarrow$} \\
        \hline
        Textual Inversion\cite{gal2022image} & 0.6533 & 0.2134 & 0.7424 & 7610 \\
        DreamBooth\cite{ruiz2023dreambooth} & 0.5518 & 0.2411 & 0.8155 & 975 \\
        Celeb Basis\cite{yuan2023inserting} & 0.6689 & 0.2278 & 0.8000 & 485 \\
        PuLID\cite{guo2024pulid} & 0.4849 & 0.1856 & 0.8039 & - \\
        \hline
         \cellcolor[HTML]{EFEFEF}\textbf{Ours (ID-EA)} &  \cellcolor[HTML]{EFEFEF}\textbf{0.6763} &  \cellcolor[HTML]{EFEFEF}\textbf{0.2427} &  \cellcolor[HTML]{EFEFEF}\textbf{0.8190} &  \cellcolor[HTML]{EFEFEF}\textbf{483} \\
        \hline
    \end{tabular}
    \vspace{-10pt}
    \label{table-jin1}
\end{table}

\textbf{Quantitative Evaluation.} 
We quantitatively evaluate the approaches from two perspectives: the identity similarity between generated images and input images and the prompt similarity between generated images and the provided text prompts. All methods are evaluated using 20 different text prompts covering background modification, interaction with other individuals, artistic styles, and expression editing. For each method, comparisons are performed by generating 32 images per prompt using identical random seeds.

The results are shown in Table~\ref{table-jin1}. DreamBooth~\cite{ruiz2023dreambooth} excels in prompt similarity but often neglects the personalized identity, whereas Textual Inversion~\cite{gal2022image} achieves high identity similarity scores but lower prompt similarity, reflecting its tendency to overfit the given input images. Celeb Basis~\cite{yuan2023inserting}, leveraging FRNet, attains strong identity preservation but exhibits lower prompt fidelity despite effective utilization of the textual latent space. PuLID~\cite{guo2024pulid} utilizes pre-trained generative models to preserve image identities without subject-specific fine-tuning. However, it often struggles to generate recognizable identities aligned with textual prompts, resulting in low identity and prompt similarity metrics. Additionally, despite avoiding per-image fine-tuning, constructing its pre-trained model remains computationally expensive and challenging to refine for improved fidelity. Our ID-EA consistently yields detailed and accurate representations of the reference image’s facial attributes while effectively aligning with the textual prompts, leading to the highest scores in all metrics. Moreover, since ID-EA requires training only a minimal set of additional parameters while preserving the pre-trained generative model’s capabilities, it achieves superior training efficiency compared to other methods.

\subsection{Ablation studies}

\begin{figure}[t] 
  \centerline{\includegraphics[width=0.48\textwidth]{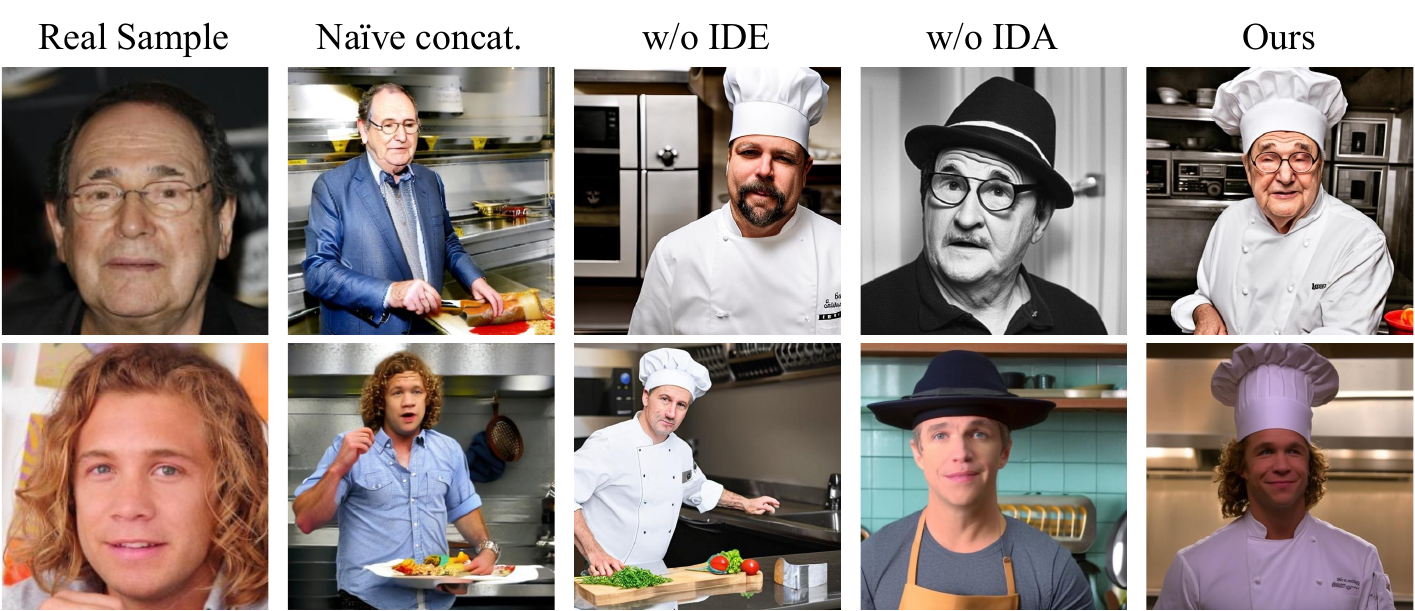}}
  \caption{
    Ablation study. The prompt is ``\( S^* \) wears a chef's hat in the kitchen". We compare the models trained with FRNet and text embeddings na\"\i ve concat. (Na\"\i ve concat.), without ID-Enhancer (w/o IDE), and without ID-Adapter (w/o IDA).
  } 
  \label{fig6} 

\end{figure}

We conduct an ablation study by sequentially introducing each submodule to evaluate their individual contributions, as shown in Fig.~\ref{fig6}. Specifically, we evaluate the following components: FRNet with simple concatenation of text embeddings, ID-Enhancer, and ID-Adapter.
In particular, in the variant with a naïve concatenation of ID-conditioned embeddings and FRNet, the model fails to properly incorporate the prompt information, resulting in images that do not reflect the intended textual description. When ID-Enhancer is removed, the model tends to discard identity information, focusing solely on the prompt tokens and neglecting the individual’s distinctive features. Without ID-Adapter, our method exhibits reduced editability, struggling to form a coherent scene as described in the prompt. As can be seen, all sub-modules are essential for achieving identity-preserved and prompt-aligned personalized face generation.

\begin{table}[t!]
\caption{Analysis of contributions of each component on ID-EA.} 
    \centering
    \renewcommand{\arraystretch}{1.5} 
    \setlength{\tabcolsep}{7.5pt} 
    \begin{tabular}{lcccc}
        \hline
        \textbf{Methods} & \textbf{Identity $\uparrow$} & \textbf{Prompt $\uparrow$} & \textbf{IQA $\uparrow$} & \textbf{Time $\downarrow$} \\
        \hline
        na\"\i ve concat. & 0.6429 & 0.1822 & 0.8079 & \underline{472} \\
        w/o IDE & 0.4868 & 0.2267 & 0.7858 & 478 \\
        w/o IDA & 0.6758 & 0.2038 & 0.7050 & 479 \\
        \hline
         \cellcolor[HTML]{EFEFEF}\textbf{Ours (ID-EA)} &  \cellcolor[HTML]{EFEFEF}\textbf{0.6763} &  \cellcolor[HTML]{EFEFEF}\textbf{0.2427} &  \cellcolor[HTML]{EFEFEF}\textbf{0.8190} &  \cellcolor[HTML]{EFEFEF}483 \\
        \hline
    \end{tabular}
    \vspace{-12pt}
    \label{table-jin2}
\end{table}

\section{CONCLUSION}
Our approach introduces a new identity enhancement and text adaptation method for personalized text-to-image generation. We identified a significant disparity between the image and prompt embeddings in text-to-image generation methods. Our ID-EA addresses this challenge by adapting the diffusion model with ID-Enhancer and ID-Adapter, achieving more identity-preserved, prompt-aligned, and faster face personalization. Our ID-Enhancer integrates textual features into image features, thus embedding textual information within visual representations. Our ID-Adapter, recombines the encoded visual information conditioned on textual contexts. While our current approach focuses primarily on generating images of human-specific concepts, we plan to extend this research to more general concepts in future work.


\bibliographystyle{IEEEtran}
\bibliography{REFERENCE}

\end{document}